\documentclass{IOS-Book-Article}

\usepackage{mathptmx}
\usepackage{soul}\setuldepth{article}
\usepackage{bbm}
\usepackage{subcaption}
\usepackage{parskip}
\usepackage{hyperref}
\usepackage{pifont}
\usepackage{amsfonts}
\usepackage{amssymb}
\usepackage{longtable}
\usepackage{xcolor}
\usepackage{multicol}
\usepackage{placeins}
\usepackage{graphicx}
\usepackage{enumitem}
\usepackage{float}
\def\hb{\hbox to 11.5 cm{}}
\begin{document}

\def\thepage{}
\begin{frontmatter}

\title{The Impact of AI Usage and Informativeness on Skill Development in Logical Reasoning
\thanks{In Hybrid Human Artificial Intelligence (HHAI) 2026, Brussels, Belgium, July 8--10, 2026. \url{https://hhai-conference.org/2026/}}
}

\author[A]{\fnms{Shang} \snm{Wu} 
\thanks{Corresponding Author: Shang Wu, University of California, Irvine, CA, 92697; E-mail:shangw13@uci.edu}},
\author[B]{\fnms{Hongyu} \snm{Yao}
},
\author[A]{\fnms{Catarina} \snm{Bel\'em}
},
\author[A]{\fnms{Shuyuan} \snm{Fu}
},
\author[A]{\fnms{Mark} \snm{Steyvers}
},
\author[A]{\fnms{Padhraic} \snm{Smyth}
}

\runningauthor{Wu et al.}
\address[A]{University of California, Irvine}
\address[B]{Massachusetts Institute of Technology}

\begin{abstract}
Artificial intelligence (AI) is being increasingly integrated into human problem-solving, yet its effects on individual skill development remain unclear. We examine how both AI usage and informativeness can shape learning in the context of a controlled logical reasoning task with on-demand access to AI assistance. We find that greater AI usage is associated with weaker skill development: heavy AI users underperform relative to comparable peers, whereas light AI users perform similarly to matched users who do not use AI. We also find in our study that these patterns are mediated by AI informativeness. Low-information AI neither improves immediate performance nor preserves performance after AI assistance is removed, and is linked to weaker learning overall. On the other hand, high-information AI  was found to improve short-run performance without reducing post-AI outcomes on average in our experiments, but with heterogeneous effects.
Our findings in general suggest that AI can, depending on context, either complement human skill development by amplifying independent reasoning or can act as a substitute that undermines such reasoning, with the implication that regulating AI access and usage will be important for promoting skill development in the presence of AI assistance.
\end{abstract}

\end{frontmatter}
\markboth{February 2026\hb}{February 2026\hb}

\section{Introduction}
Artificial intelligence (AI) assistance is being increasingly adopted in our everyday lives across a variety of contexts such as education \cite{noy2023experimental, kasneci2023chatgpt},  software development \cite{peng2023impact,vaithilingam2022expectation}, autonomous driving \cite{kalra2016driving}, and medical diagnosis \cite{budzyn2025endoscopist}.
This has led to increased recent research attention on topics such as human-AI complementarity \cite{steyvers2022bayesian, wilder2020learning, fugener2026roles}. In parallel, there are concerns that easy access to AI may foster overreliance, reduce independent effort, and potentially hinder learning from the perspective of human users \cite{gerlich2025ai, macnamara2024does}. However, there have been relatively few studies to date that directly examine how AI assistance affects individual skill development over time. In this paper, we study this issue with a particular focus on assessing skill development after AI assistance has been removed. 

The literature to date on the impact of AI assistance has provided a number of alternative perspectives on the role of AI assistance in the context of human skill learning. Some studies argue that richer AI assistance, such as detailed guidance or uncertainty estimates, can improve human performance \cite{zhang2020effect, senoner2024explainable}. Other studies caution that limiting AI assistance may help reduce human overreliance and promote independent thinking \cite{buccinca2021trust,de2025cognitive}. Moreover, most prior work focuses on immediate performance gain rather than comparing pre-AI and post-AI learning outcomes \cite{buccinca2021trust, lai2019human, brynjolfsson2025generative}. To provide a more nuanced understanding, we investigate how both (i) the amount of AI usage, and (ii) the amount of information provided by AI, shape the trajectory of individual skill development.

We study how access to AI assistance of varying informativeness shapes individual skill development over time. Specifically, we address several related questions. First, we ask whether greater AI usage, as a behavioral pattern of reliance, is associated with weaker subsequent latent skill development. We then investigate how the informativeness of AI assistance moderates this relationship. Specifically, we examine whether the amount of information provided by the AI differentially alters immediate performance and subsequent latent skill development. Finally, we examine how individuals' AI usage strategy drives the differences in observed learning outcomes. 

\begin{figure}[tb] 
  \begin{center}
    \includegraphics[width=0.80\textwidth]{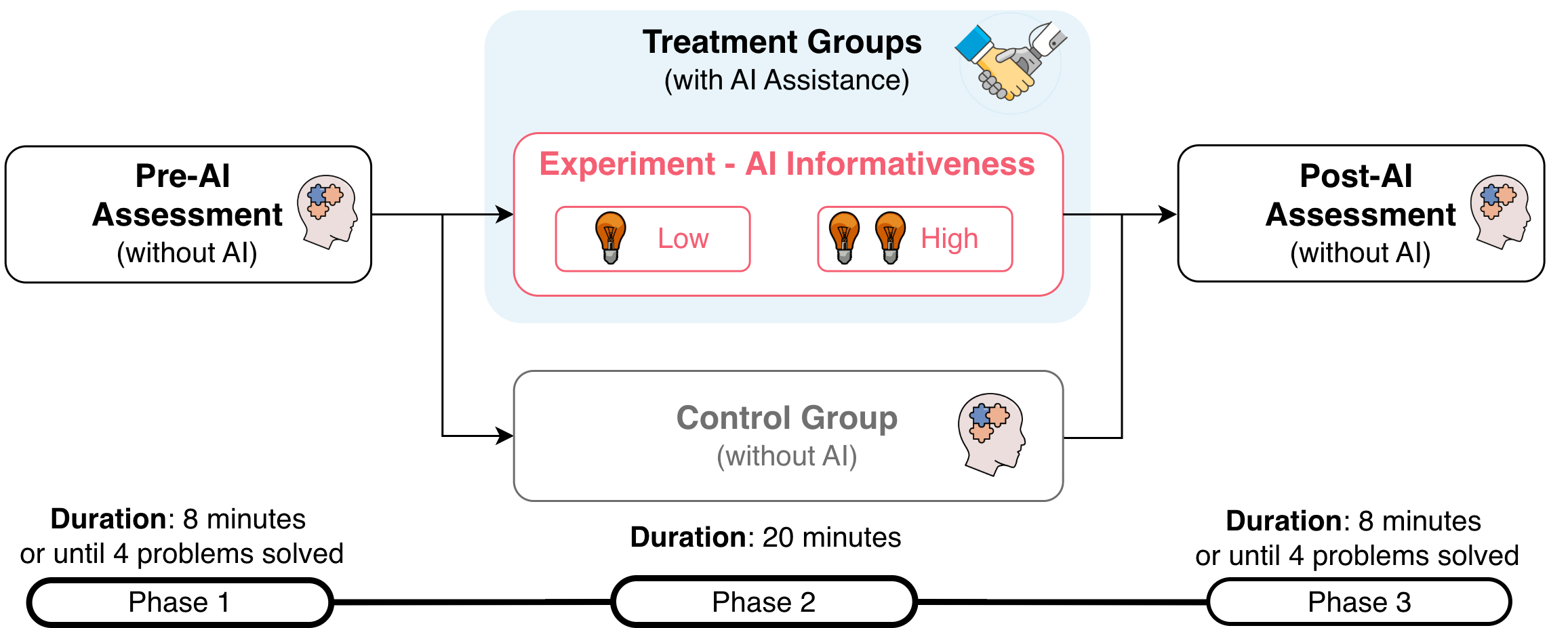}
  \end{center}
\caption{
Study design with three phases: Phase~1 (pre-AI) and Phase~3 (post-AI) human performance was assessed without any assistance. In Phase~2, participants were assigned to either a control group (no AI) or a treatment group with optional AI assistance that varies in AI informativeness.
}
\label{fig:exp_diagram_flow}
\end{figure}

We investigate these questions through a controlled user study with time-constrained logical reasoning tasks. Participants solve puzzles under conditions that allow on-demand interaction with a simulated AI system. 
Each participant completes assessments both before and after AI exposure without any assistance, enabling us to isolate the effect (of interacting with AI) on skill development beyond the period of AI use (see Figure \ref{fig:exp_diagram_flow}). The experimental design systematically manipulates the level of information provided by the AI, a central treatment dimension aligned with our theoretical focus on reliance and learning. Our goal is to understand how varying the informativeness of assistance changes AI usage behavior and, in turn, affects individual learning outcomes.

Our results show that AI assistance is associated with weaker skill development in logic-based problem-solving tasks. This effect is driven by usage behavior: earlier and more frequent reliance on AI leads to less independent problem-solving effort. As a result, heavy AI users underperform comparable peers in terms of skill development, whereas light AI users slightly outperform peers with similar initial ability who did not use AI. The magnitude and direction of these associations vary with AI informativeness. High-information AI substantially improves immediate performance and, on average, does not appear to impair skill growth. In contrast, low-information AI neither meaningfully improves immediate performance nor preserves subsequent performance. Moreover, high-information AI is linked to heterogeneous effects: higher-ability individuals tend to use it more selectively and later in the problem-solving process, and exhibit stronger skill growth. On the other hand, lower-ability individuals rely more heavily and earlier, and show comparatively weaker learning outcomes. Lower-ability participants report elevated perceived ability when AI is available, suggesting a possible role for miscalibrated self-assessment that may, in turn, be associated with weaker subsequent learning.

Collectively, these findings underscore that the educational and productivity consequences of AI depend not only on access but also on its informativeness and user heterogeneity. AI can augment learning when it complements sustained cognitive engagement, but may hinder skill development when it substitutes for it, particularly among individuals most inclined to rely on it.

\section{Related Work}
The concept of cognitive skill development refers to durable improvements in reasoning capacity, problem-solving ability, and learning transfer beyond immediate task performance \cite{vygotsky1978mind}. 
In this context, a central question in recent human–AI interaction (HAI) research is whether AI assistance in general merely affects immediate performance or can also meaningfully affect longer-term underlying skill development (either positively or negatively). Research in HAI has largely emphasized human-AI complementarity and performance gains during AI use \cite{steyvers2022bayesian, wilder2020learning}. There has been less research on the evaluation of downstream learning outcomes, and the results of that work have been mixed in their conclusions. Some studies show that external assistance can scaffold reasoning and improve retention when appropriately designed \cite{yan2025effects, gajos2022people}. Related work suggests that AI is beneficial when it complements rather than substitutes for human thinking \cite{poulidis2025action}. In contrast, other work documents reduced cognitive effort, increased offloading, and weaker skill development \cite{kosmyna2025your, shen2026ai}. Additional evidence suggests that AI assistance can improve immediate task performance without inducing skill degradation \cite{karny2024learning}. These studies reveal a consistent tension between immediate support and long-term skill development. In addition, much of the previous literature has had only limited emphasis on evaluating human performance after AI assistance is removed, making it difficult to distinguish genuine skill development from temporary reliance on external support. 
Pre- and post-assessment designs provide a clear approach by measuring baseline ability before AI exposure and reassessing performance after the AI is removed \cite{dimitrov2003pretest}, making it possible to attribute changes to learning rather than to ongoing assistance.

Another underexplored dimension concerns individual heterogeneity. Prior work shows that reliance behaviors, such as help-seeking or AI adoption rate, vary substantially across individuals \cite{gu2025personalized, swaroop2024accuracy}. However, many studies either provide continuous AI access with limited user autonomy \cite{cao2023time, fogliato2022goes} or rely on self-reported measures of traits and trust \cite{swaroop2024accuracy, lee2025impact}, which are vulnerable to reporting biases \cite{joinson1999social}. Moreover, relatively little work systematically connects objectively measured baseline ability to subsequent reliance patterns and post-assistance learning outcomes. As a result, it remains unclear whether AI engagement differences reflect stable individual skill differences or context-specific behavioral responses.

Additionally, prior research shows that design features such as timing and explanations displays influence reliance and performance \cite{karny2024learning,swaroop2024accuracy, vasconcelos2023explanations}, and some studies vary AI accuracy to study these effects \cite{tejeda2023displaying}. However, manipulating accuracy introduces trust dynamics: early errors can shift beliefs and confound perceived reliability with the intrinsic AI assistance value \cite{kahr2024trust}. Moreover, perspectives diverge on the role of informational depth. Some argue that richer guidance enhances understanding and performance \cite{zhang2020effect, senoner2024explainable}, while others suggest that limited assistance encourages deeper engagement and mitigates overreliance \cite{buccinca2021trust, de2025cognitive}. Yet few studies cleanly isolate the effect of AI informativeness itself by assessing skill levels after AI assistance is removed.

We address these gaps by holding AI accuracy constant while experimentally varying informational depth. Combined with a pre- and post-assessment design and an objective measure of baseline ability, our framework allows us to examine how initial ability influences AI reliance and subsequent skill development, as well as how differences in AI informativeness affect learning once the AI is removed.

\section{Experiment}
We conducted a controlled user experiment to study how AI assistance affects individual skill development when help-seeking is self-directed, and tasks have time constraints. We manipulated the level of AI informativeness to examine how AI engagement influences learning and reliance.

\subsection{Task Description}
We designed a time-constrained, logic-based puzzle in which participants determine the unique ordering of six objects subject to a set of constraints. Although this type of structured logic puzzle is an abstraction of real-world situations, it nonetheless offers a controlled environment that isolates the impact of AI assistance on problem-solving and learning.

\begin{figure}[tb]
\centering

\begin{subfigure}{0.05\textwidth}
\textcolor{white}{\%}
\end{subfigure} %
\begin{subfigure}{0.50\textwidth}
    \raggedleft
    \includegraphics[height=0.17\textheight]{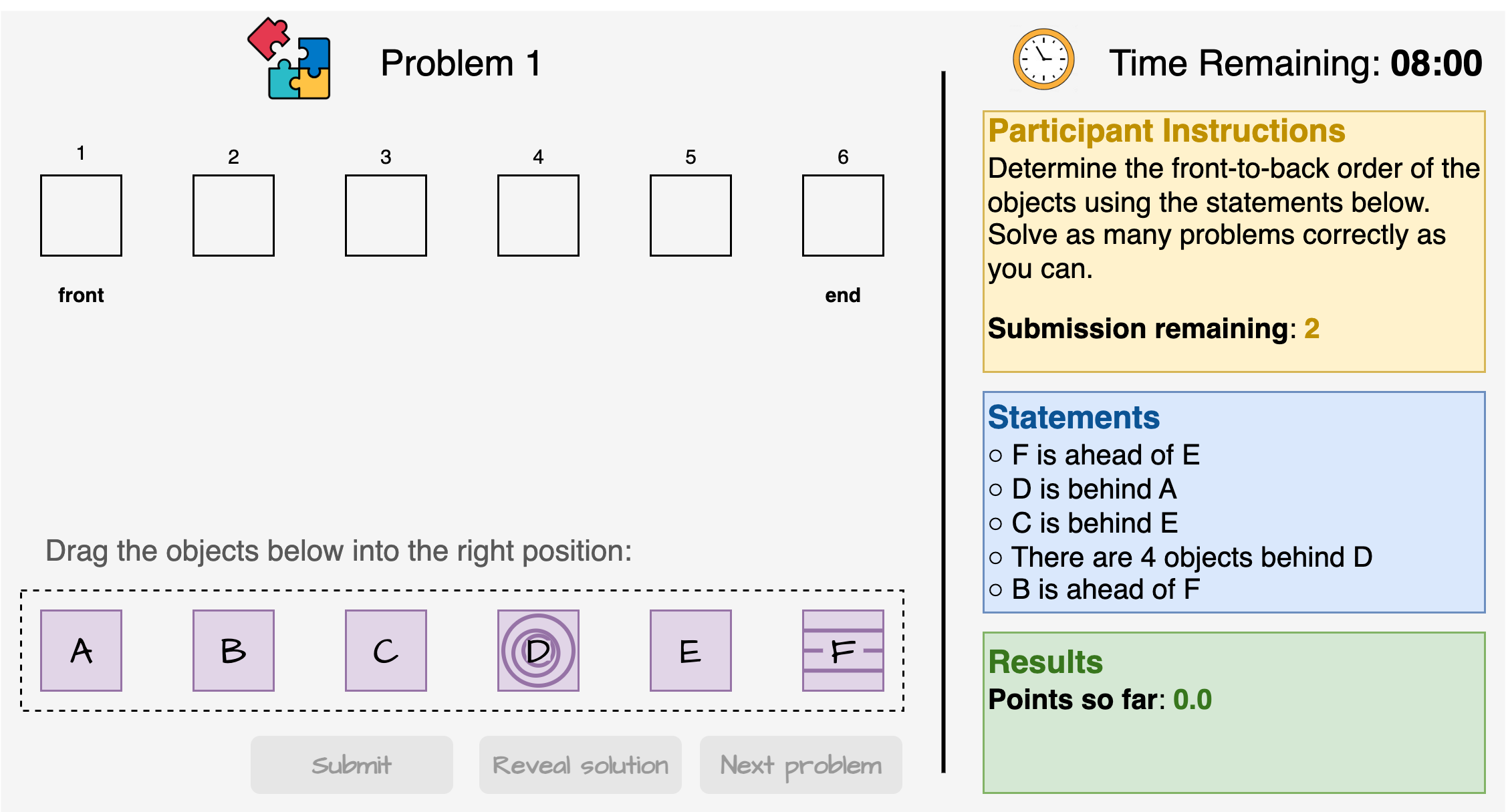}
    \caption{Without AI assistance}
    \label{fig:screenshot_noAI}
\end{subfigure}
\hfill
\begin{subfigure}{0.6\textwidth}
    \raggedleft
    \includegraphics[width=\linewidth]{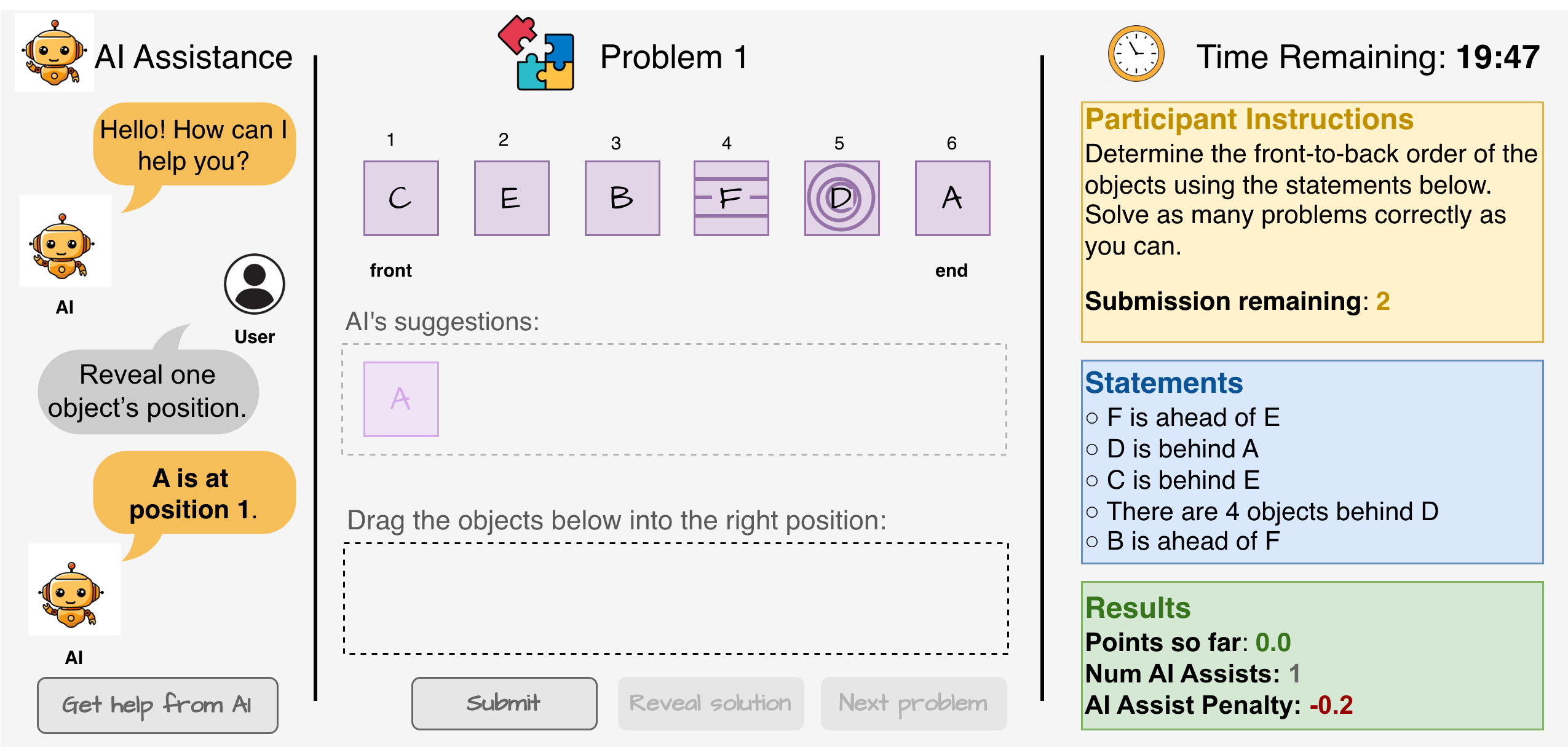}
    \caption{With AI assistance}
    \label{fig:screenshot_withAI}
\end{subfigure}
\caption{Examples of user interfaces for the logic-based puzzle task}
\end{figure}

For each problem, participants were presented with six randomly ordered objects along with several logical statements that jointly implied a correct and unique sequential ordering (Figure~\ref{fig:screenshot_noAI}). Participants could submit up to two proposed solution orderings per problem. 
After the first submission, the user is given feedback to let them know how many objects are in their correct location in the sequence.   
If all six objects are in the correct order, the user proceeds to the next problem. If one or more objects are not in the correct position, the participants can either (i) accept their initial attempt as final or (ii) revise their answer and submit the revision. In all cases, the user is always shown the correct solution before moving to the next problem, in order to provide feedback to support learning across problems. Additionally, certain objects were marked with hints linked to specific positions, for example, an object with a horizontal line was more likely to be placed in Position 4 (see Figure~\ref{fig:screenshot_noAI}, object F), thereby facilitating learning and knowledge transfer.

The study consisted of three phases (Figure \ref{fig:exp_diagram_flow}). Phase 1 functioned as a pre-AI assessment and lasted 8 minutes, with participants required to complete at least four problems; those who had not completed four within 8 minutes were allowed to continue until they did. Phase 2 (20 minutes) introduced the experimental conditions, including optional AI assistance, as described in the Conditions section below. Phase 3 replicated Phase 1 as a post-AI assessment. Problem order was randomized within each phase to control for difficulty-related learning effects.
Participants earned one point per correctly solved problem. 

\subsection{Conditions}
The study used a between-subjects design with three Phase 2 conditions: \textit{No-AI}, \textit{Low-information AI}, and \textit{High-information AI}. AI assistance was available exclusively in Phase 2 and could be requested at most once per problem, allowing participants to decide whether and when to seek assistance. Upon requesting help, the AI assistant revealed the positions of randomly selected objects, with the number of revealed objects determined by the assigned condition. Each AI request incurred a 0.2-point deduction, applied only if the problem was solved correctly. Participants were informed of this cost in advance.

Participants were randomly assigned to one of the following conditions:
\begin{itemize}[noitemsep, topsep=0pt]
\item \textit{\textbf{No-AI}}: No AI assistance was available to participants (see Figure~\ref{fig:screenshot_noAI}).
\item \textit{\textbf{Low-information AI}}: The AI can randomly reveal, on request during Phase 2, the location of one object (see Figure \ref{fig:screenshot_withAI}).
\item \textit{\textbf{High-information AI}}: The AI can randomly reveal, on request during Phase 2, the locations of three objects.
\end{itemize}

To rule out AI accuracy as a potential confound when examining how informativeness affects human reliance and performance, we used a simulated AI agent with perfect (100\%) accuracy. Simulated AI agents are commonly employed in human–AI interaction research (e.g., \cite{panigutti2022understanding, srivastava2022improving, cao2024designing, eisbach2023optimizing}), particularly in settings where AI is assumed to be fully accurate to isolate specific behavioral mechanisms \cite{panigutti2022understanding, srivastava2022improving}. Participants were not told about the AI’s accuracy and were informed only that AI assistance would be available during Phase 2 at a cost.

\subsection{Procedure}
The study was conducted online via Prolific. After providing informed consent, participants were randomly assigned to an experimental condition and presented with instructions. They were required to pass two comprehension checks before moving on, and only those participants who answered both correctly proceeded to the main task. The study consisted of three phases, separated by brief screen breaks. An attention check was embedded in Phase~2 to maintain data quality. Upon completion of the final phase, participants filled out a short post-study survey.

\subsection{Participants}
We recruited 160 U.S.-based, English-speaking adults (18+) with at least an undergraduate degree. After excluding 28 participants due to attention failures, inattentiveness, inconsistent AI-use reports, and extreme outliers, the final sample comprised 132 participants (42 \textit{No-AI}, 43 \textit{High-information AI}, 47 \textit{Low-information AI}). 
Participants had a mean age of 39.7 (SD = 11.9); 58 identified as male, and 74 as female; 67 had an undergraduate degree and 65 had a master’s degree or higher.

Participants received \$9 for participation (median duration: 60 minutes) and could earn up to \$3.50 in performance-based bonuses. The experiment received Institutional Review Board (IRB) approval at the University of California, Irvine, protocol number \#7206.

\subsection{Measures}
\label{subsection:exp1_measure}
We report below seven primary metrics from our study: 

\begin{itemize}[noitemsep, topsep=0pt]
    \item \textbf{\textit{Response Time}}: Time spent on each problem (seconds) from when the participant first sees the problem to when they submit their final solution and proceed to the next problem.
    \item \textbf{\textit{Correctness}}: Number of objects placed in the correct position (0–6). 
    \item \textbf{\textit{Reward Rate}}: Correctness divided by response time (converted to minutes), capturing the speed–accuracy trade-off in cognitive modeling \cite{standage2015toward}. Higher values indicate better performance.
    \item \textbf{\textit{Initial Ability}}: Participants’ Phase 1 reward rate. We further split participants into low- and high-ability groups based on the median of initial ability.
    
    \item \textbf{\textit{Timing of Request for AI Assistance}}: 
    Time from problem start to the first AI request, conditional on AI assistance being requested.

    \item \textbf{\textit{AI Usage Fraction}}: The percentage of Phase~2 problems in which participants requested AI assistance. The upper bound for this metric is 1.0, indicating that a participant used AI assistance in every problem in Phase~2.
    
    \item \textbf{\textit{Solo Thinking Ratio}}: 
    From a cognitive offloading perspective, one observable marker of engagement is the amount of time individuals devote to independent problem-solving before seeking external assistance. We use the ratio of \textit{solo thinking time} to response time, where \textit{solo thinking time} refers to the period participants worked independently before requesting AI assistance (i.e., same as \textit{Timing of Request for AI Assistance} above), or, if no AI was used, the entire problem-solving time until submission. We also report \textbf{\textit{Solo Share}}, defined as the participant-level average of the \textit{Solo Thinking Ratio} across all problems completed within a phase. Higher values indicate greater sustained independent problem-solving before seeking AI assistance.

\end{itemize}

The primary focus of our analysis is on reward rates at the participant level. We also examine response time and accuracy at the problem level to separate the effects of speed and accuracy. Additionally, we discuss the \textit{solo share} metric, which reflects how participants balance working independently versus seeking external help over time, allowing us to explore how individual problem-solving effort relates to skill development.

\section{Results}
\begin{figure}[h]
    \centering
        \includegraphics[width=0.65\textwidth,height=0.26\textheight,keepaspectratio]{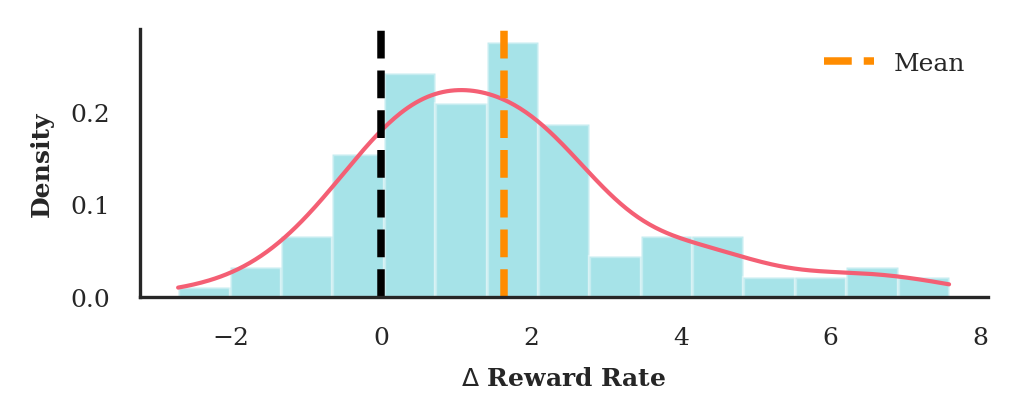}
        \caption{Distribution of individual changes in \textit{Reward Rate} (correct objects per minute), defined as Phase~3 $-$ Phase~1. The black dashed line marks zero-change reference, and the colored dashed line denotes the sample mean.}
      
    \label{fig:delta_learning_effect}
\end{figure}

We begin by providing a high-level summary of skill development trajectories across individuals over the duration of the experiment. Figure~\ref{fig:delta_learning_effect} displays the distribution of individual-level changes in reward rate from Phase~1 to Phase~3. Most participants exhibit gains in skill (reward rate) over time, and the average improvement is statistically significant (mean improvement = 1.64, $p<0.01$), providing evidence of overall skill development over time (the median reward rate increases from 1.89 in Phase~1 to 3.35 in Phase~3). The substantial spread of the distribution reveals substantial cross-individual variation in the magnitude of these learning gains.

\FloatBarrier

\FloatBarrier
\begin{figure}[h]
\centering
\begin{subfigure}[t]{0.48\textwidth}
    \centering
    \includegraphics[width=\textwidth]{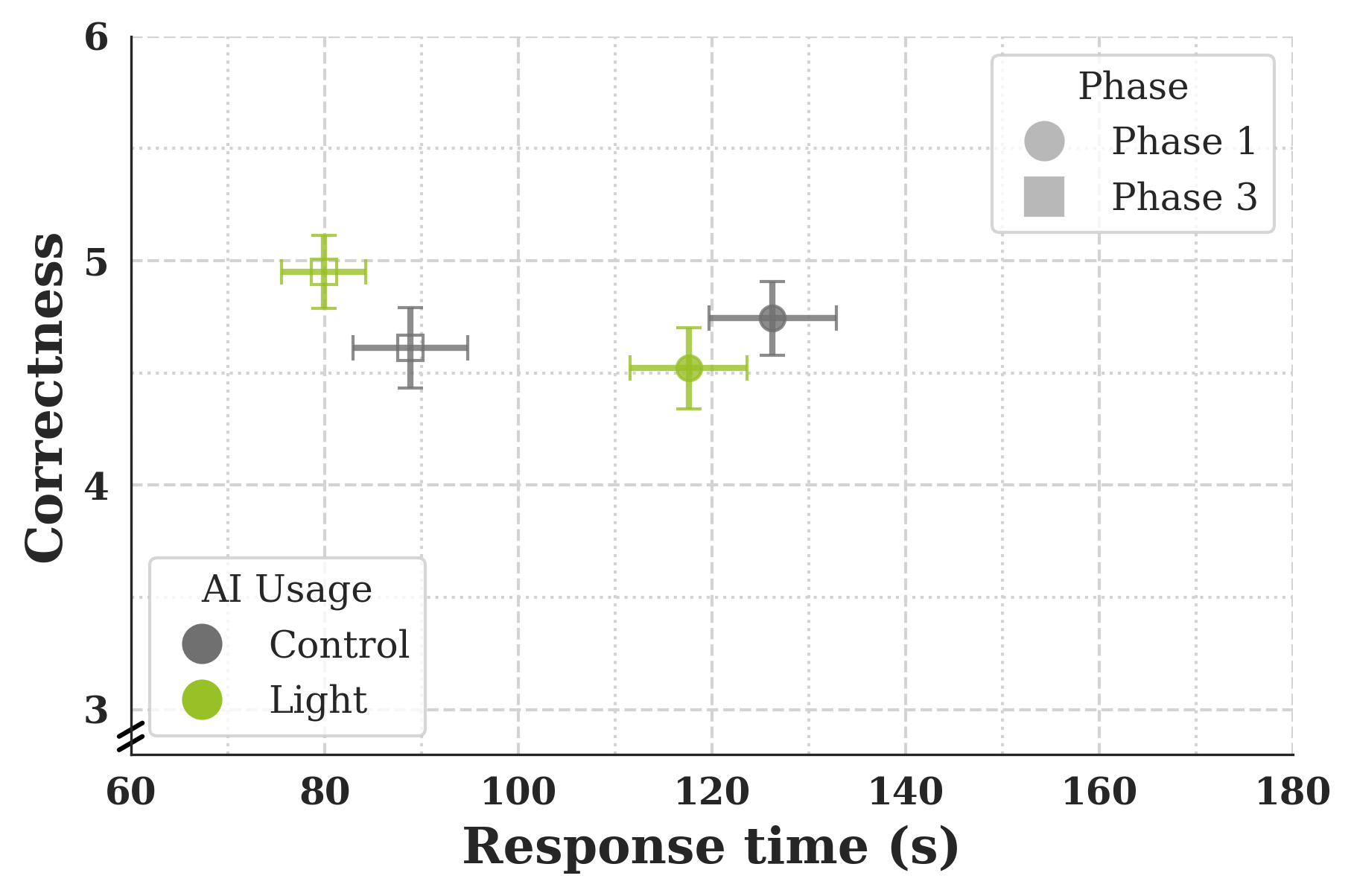}
    \caption{Light usage group vs. matched control group}
    \label{fig:psm_light}
\end{subfigure}
\hfill
\begin{subfigure}[t]{0.48\textwidth}
    \centering
    \includegraphics[width=\textwidth]{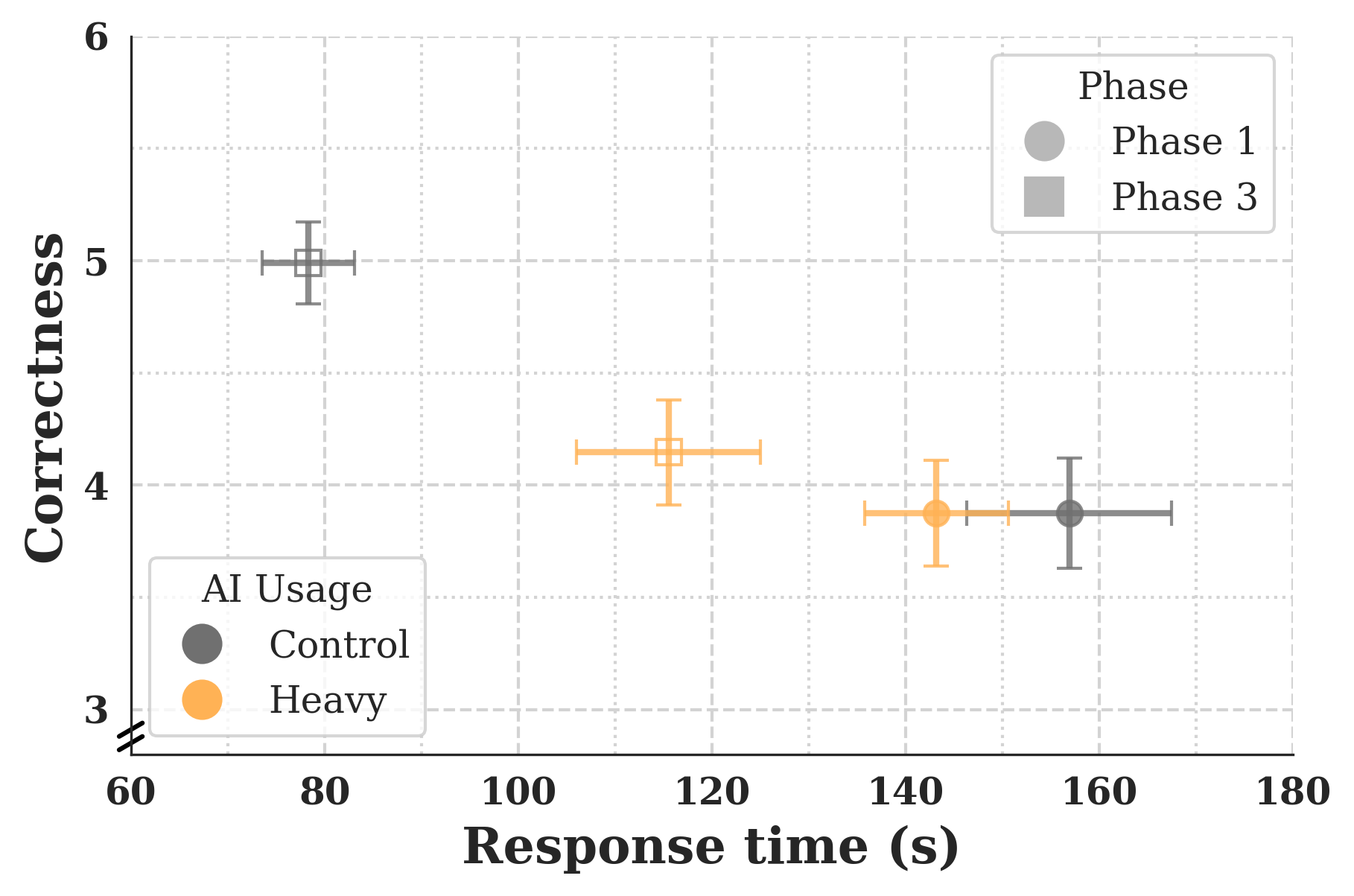}
\caption{Heavy usage group vs. matched control group}
\label{fig:psm_heavy}
\end{subfigure}
\caption{Propensity score matching (PSM): average correctness and response time (with standard error bars) for light and heavy AI usage groups versus matched control groups in Phase~1 (pre-AI) and Phase~3 (post-AI).}
\label{fig:psm}
\end{figure}

\subsection{When AI assistance hinders skill development}
We start by analyzing the relationship between individual AI usage and subsequent skill development. Participants are categorized according to their AI usage fraction (the fraction of Phase 2 problems where participants requested AI assistance): Zero (no AI requests), Light (usage fraction $\in (0, 0.4]$), and Heavy (usage fraction $\in (0.4, 1]$), where 0.4 corresponds to the mean usage rate among those who requested AI at least once. 
To account for differences in participants’ initial ability, we conduct propensity score matching (PSM) \cite{austin2011introduction}, matching Light and Heavy users separately to Zero-usage participants based on Phase~1 performance in the two-dimensional space of correctness and response time, using logistic regression with nearest-neighbor matching. Each light-AI user ($N=36$) is matched with one control, and each heavy-AI user ($N=24$) with one control. Results are robust to alternative thresholds, such as a median split.
Figure \ref{fig:psm} represents the results. In Phase~1, treatment and control groups do not differ significantly, as expected, since matching was based on initial performance. In Phase~3, however, relative to the matched controls, heavy AI users have significantly longer response time (115.54 (se = 9.49) vs. 78.31 (se = 4.76); $p< 0.01$) and lower correctness (4.15 (se = 0.23) vs. 4.99 (se = 0.18); $p< 0.01$), suggesting weaker individual skill development. Light AI users are statistically indistinguishable from their matched controls in response time, exhibiting slightly higher correctness (4.95, se = 0.16 vs. 4.61, se = 0.18; $p=0.08$), suggesting that light AI use does not slow participants down and may modestly improve accuracy.
These patterns strongly indicate, in the context of logic puzzle-solving, that greater reliance on AI is associated with weaker learning, whereas limited and selective use may be compatible with, or even supportive of, skill growth.

\subsection{AI informativeness moderates immediate performance and learning outcomes}

We next discuss how these usage patterns interact with AI informativeness. Figure \ref{fig:rr_over_time} presents reward rates over time. The first and last points correspond to Phase 1 and Phase 3, respectively, while the first and second halves of Phase 2 are represented by two intermediate points.\footnote{Since Phase 2 lasts 20 minutes, we divide it into two halves to capture more granular dynamics and to make the time intervals more comparable to Phases 1 and 3.} In Phase~1, reward rates do not differ significantly across groups, indicating balance in baseline reward rate. During Phase~2, participants with access to high-information AI achieve higher reward rates ($p < 0.10$), with gains emerging immediately in the first half of Phase 2 and persisting into the second half. However, the short-run improvement under low-information AI is comparable to that observed in the control group, indicating no additional gain beyond the baseline learning. 
In Phase~3, participants in the low-information condition exhibit significantly lower reward rates than the no-AI control group ($p=0.08$), whereas the high-information group does not differ significantly from the control.
These patterns suggest that while heavier AI usage is associated with weaker learning overall, the magnitude and even the direction of these effects depend on the quality of assistance provided: high-information AI improves immediate performance without reducing average post-AI outcomes, whereas low-information AI neither enhances immediate performance relative to the control group nor preserves subsequent skill growth.

\begin{figure}[tb]
      \centering
\includegraphics[width=0.70\textwidth]{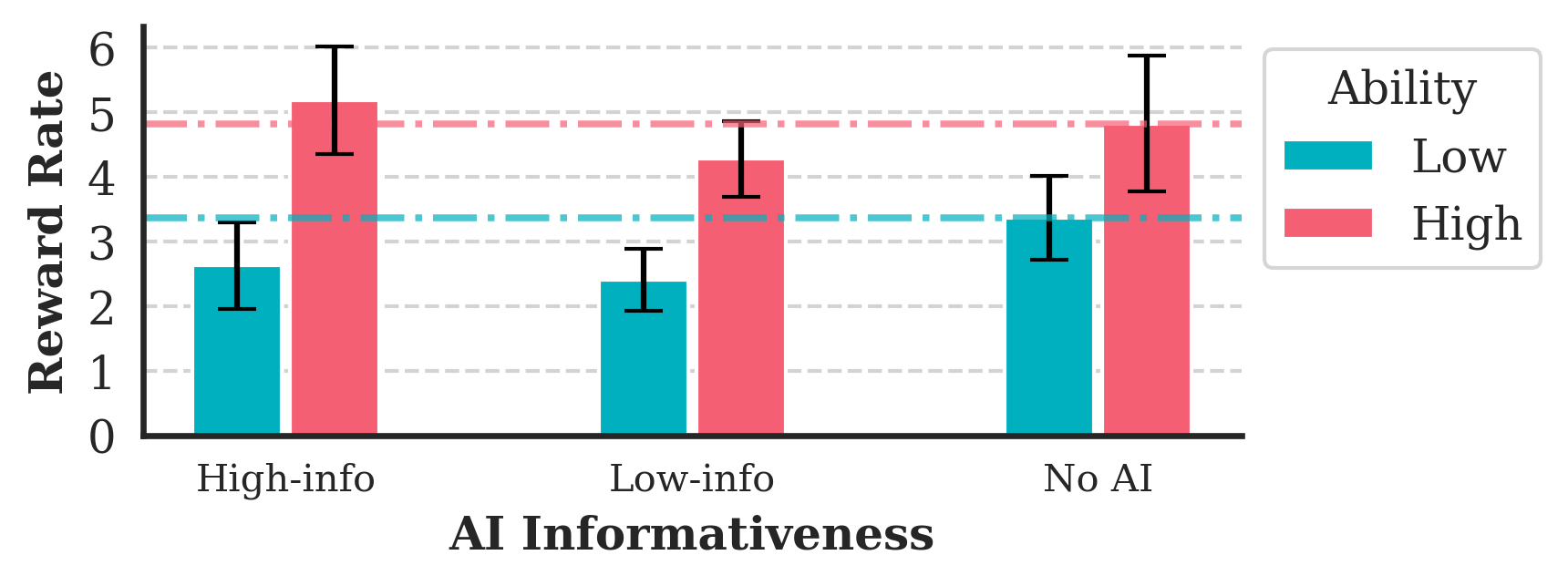}
    \caption{Post-AI average reward rates by initial ability level across different AI informativeness conditions. Error bars show 90\% confidence intervals; dashed lines show post-AI average reward rates for high- and low-ability groups in the no-AI condition.}
    \label{fig:final_rr}
\end{figure}

We further examine Phase 3 reward rates by AI informativeness and initial ability (see Figure \ref{fig:final_rr}). Two patterns emerge. First, the low-information AI condition shows a uniform decline: both high- and low-ability participants exhibit lower Phase 3 reward rates than those in the no-AI group. The decline is statistically significant for low-ability participants ($p < 0.05$) but not for high-ability participants, indicating that low-information AI is associated with broadly negative post-AI outcomes, particularly among lower-ability participants.
Second, the high-information AI condition displays a polarization effect: the post-AI performance gap between high- and low-ability participants is substantially larger under high-information AI (5.19 vs. 2.63; $p < 0.01$) than in the no-AI condition (4.82 vs. 3.36; $p = 0.06$).
That is, in the context of logic puzzles, high-information AI assistance widens the ability gap. The next two subsections explore the mechanisms behind these results.

\FloatBarrier

\begin{figure}[h]
    \centering
    \includegraphics[width=0.90\textwidth]{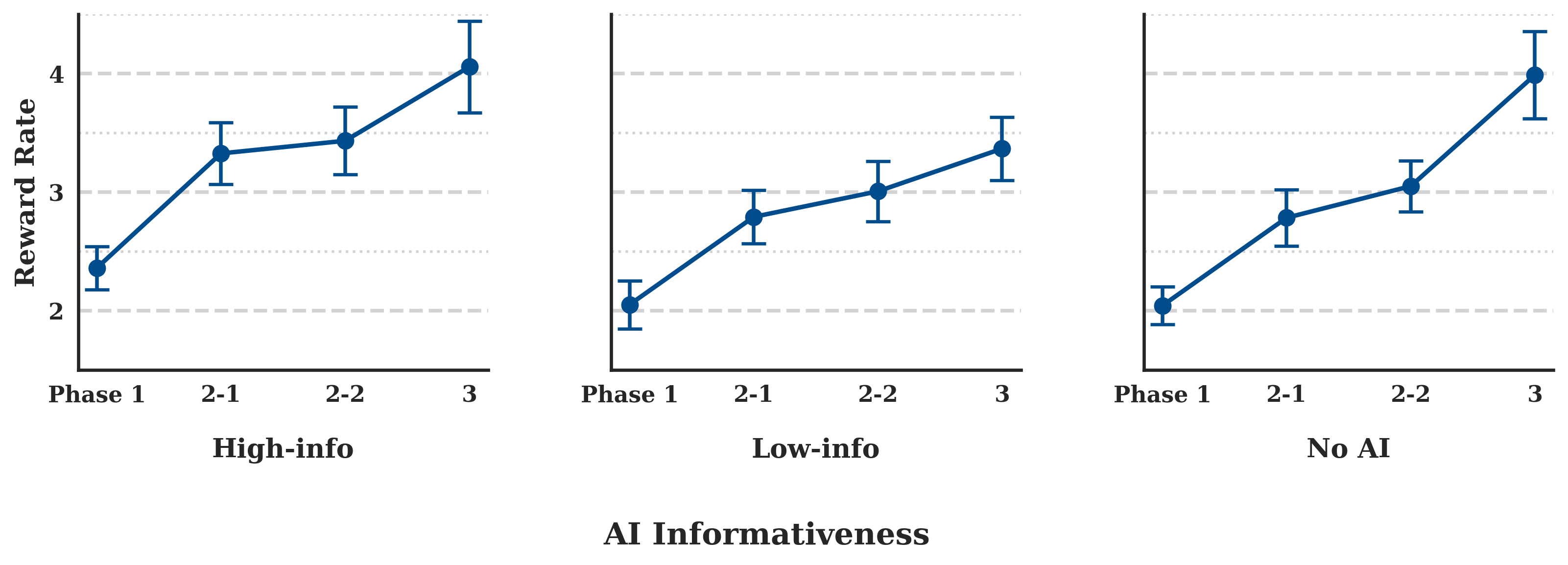}
    \label{fig:rr_over_time}
\caption{
Reward rate trajectories across experimental phases by AI informativeness. The x-axis denotes Phase 1, Phase 2–first half (2-1), Phase 2–second half (2-2), and Phase 3. Error bars represent standard errors.}
\label{fig:rr_over_time}
\end{figure}

\subsection{Low-information AI displaces independent effort without performance gains}
Here we investigate the mechanism underlying the uniform decline observed under low-information AI. Unlike the high-information condition, this version of AI provided only limited assistance: each request revealed at most one object’s location and, in some cases, repeated information already implied by the problem (e.g., the problem statement could specify that A is at position 1, and AI could then provide the same redundant suggestion). As a result, the AI had a limited effect on reducing the core complexity of the task. We conjecture that this limited assistance interrupted participants’ independent reasoning and strategy formation without meaningfully simplifying the problem. Evidence from Phase~2 behavior supports this interpretation. The average Phase 2 \textit{solo share}, which represents the proportion of independent thinking time within total response time across problems, declines from a baseline of 0.95 in the no-AI condition to 0.85 under low-information AI, and further to 0.65 for problems in which AI is actively used. This pattern indicates that even limited AI assistance is associated with a substantial reduction in independent problem-solving effort.

\begin{figure}[h]
    \centering
    \includegraphics[width=0.70\textwidth]{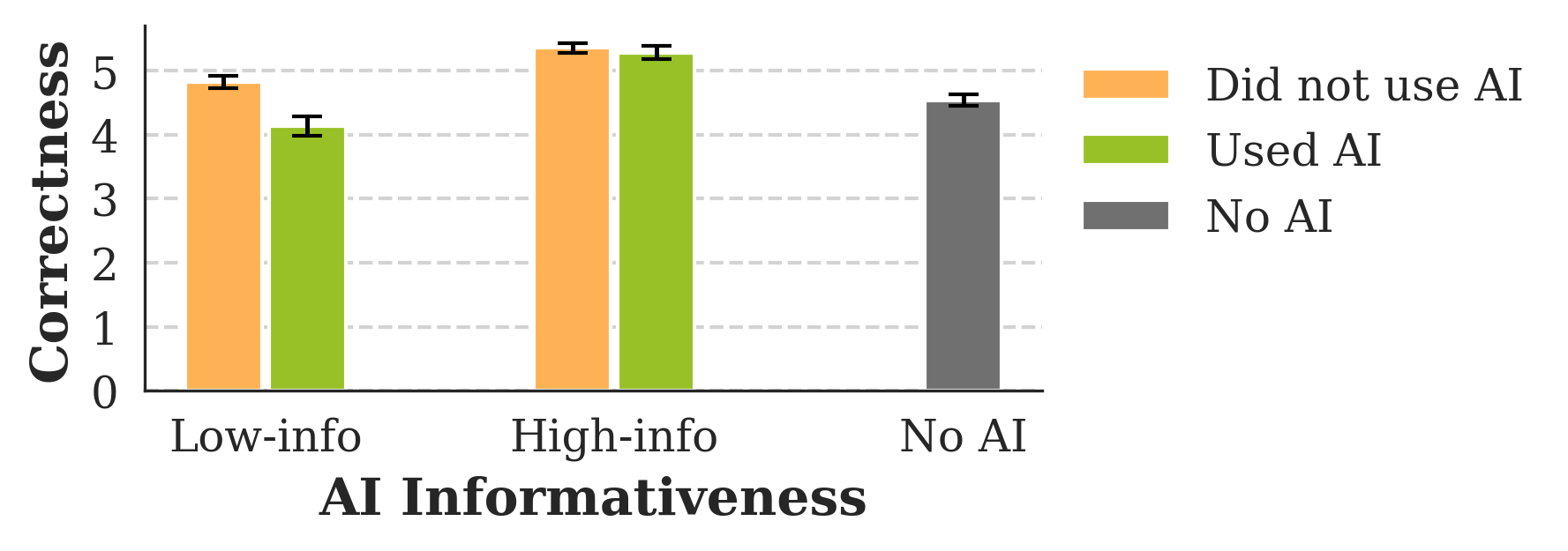}
        \caption{Phase~2 average problem-level correctness by AI informativeness and usage. Error bars represent standard errors.}
    \label{fig:correctness_by_aiinfo_usage_se}
\end{figure}

To examine whether this reduction in independent effort translates into immediate performance gains, Figure \ref{fig:correctness_by_aiinfo_usage_se} shows average correctness {\it at the problem level}, broken down by condition and AI usage for Phase~2 (where AI assistance was available). We use correctness rather than reward rate because reward rate depends on response time, which in Phase~2 may reflect interruptions to the user due to accessing AI or additional review time, rather than true performance. Focusing on the low-information condition, problems assisted by AI exhibit lower correctness than both non-assisted problems within the same group and the no-AI control, while non-assisted problems in the low-info group perform relatively better. When the high-information AI is used, correctness is significantly higher than the no-AI control. These patterns suggest that low-information AI reduces independent effort without having compensating immediate gains.

\FloatBarrier
Post-study survey responses reinforce this interpretation. Participants rate the low-information AI as significantly less helpful than high-information AI (mean 3.44 vs.\ 4.00; $p<0.05$). Nevertheless, despite recognizing its limited usefulness, participants still chose to use the low-information AI, where, on average, the AI usage fraction is 0.22. This suggests that even objectively weak AI tools can attract engagement and divert attention, making low-quality assistance potentially distracting and detrimental to effective learning.

\subsection{High-info AI induces ability-based divergence}
\begin{figure}[h]

  \centering
  \includegraphics[width=0.90\textwidth]{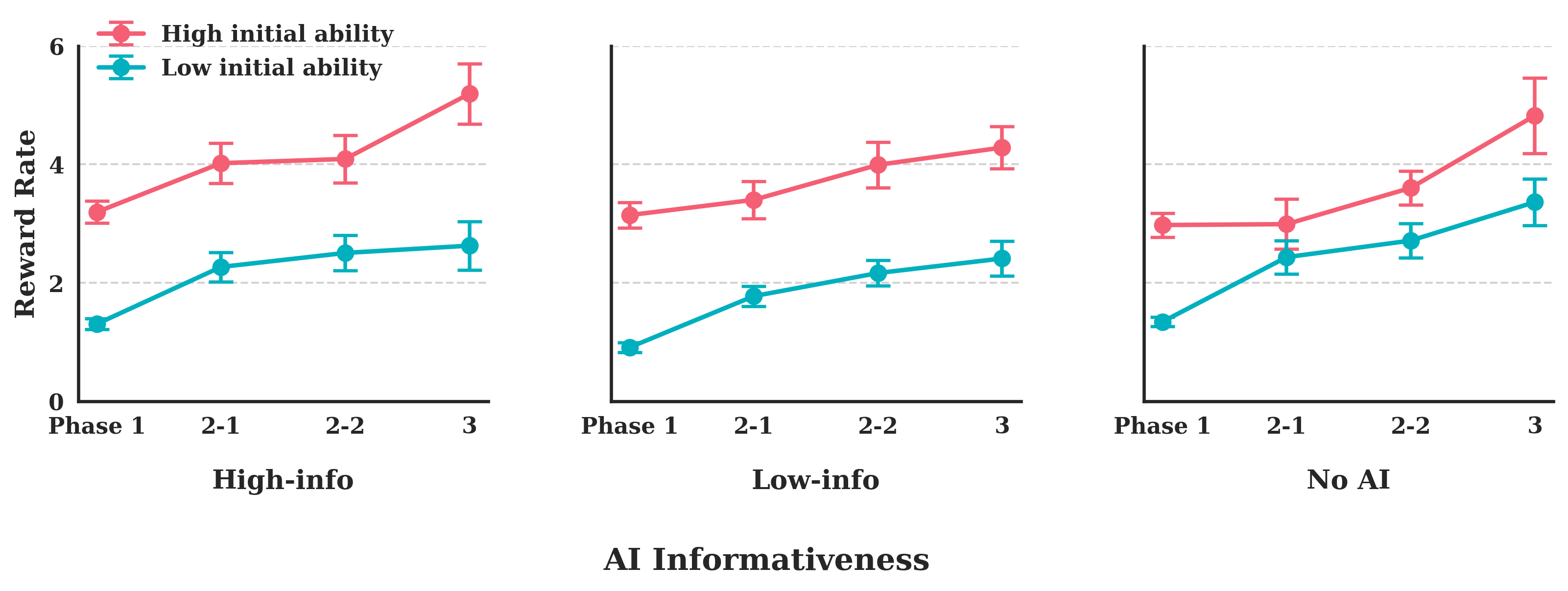}

\caption{
Reward rate trajectories across experimental phases by AI informativeness and initial ability. The x-axis denotes Phase 1, Phase 2–first half (2-1), Phase 2–second half (2-2), and Phase 3. Error bars represent standard errors.}
\label{fig:rr_over_time_by_ability}
\end{figure}

As mentioned above, high-information AI does not affect Phase~3 performance on average, but it widens the performance gap between high- and low-ability participants. Figure~\ref{fig:rr_over_time_by_ability} further shows that high-ability participants in the high-info AI condition benefit not only in long-run skill growth but also in immediate Phase 2 performance, particularly in the first half, relative to the no-AI group. In contrast, low-ability participants do not exhibit comparable short-run gains. To understand the mechanism, we examine independent problem-solving time (\textit{solo share}) and AI usage within the high-information AI condition. We find that low-ability participants spend significantly less time solving problems independently than high-ability participants (\textit{solo share} = 0.81 vs. 0.91, $p<0.01$). Low-ability participants also use AI more frequently (0.37 vs. 0.22), reflecting greater reliance. Heavier reliance seems to hinder learning: lower-ability participants tend to turn to AI earlier and use it more intensively, which ultimately reduces the potential benefits of high-quality assistance. Moreover, high-information AI appears to inflate confidence among lower-ability participants. Despite weaker objective performance, in the post-study survey, low-ability participants report a stronger belief that they have “found a strategy” (mean 4.37 vs. 4.04   for high-ability participants). In the control group, self-assessments are similar across ability levels (4.13 vs. 4.06), and high-ability participants’ ratings remain stable across conditions. This pattern suggests that high-information AI may increase the divergence between perceived and actual competence among lower-ability individuals, increasing the gap between perceived and actual competence.

\FloatBarrier

\section{Discussion}

Our findings both align with and extend prior work in human–AI interaction and the learning sciences. Consistent with research showing that excessive external support can undermine independent reasoning \cite{buccinca2021trust, wecks2024generative}, we find, in the context of our logic puzzle study, that heavy reliance on AI reduces independent problem-solving effort and is associated with worse post-AI performance.
Specifically, heavy AI users and individuals with lower initial ability exhibit the largest post-AI performance declines, echoing concerns that automation may substitute for internal cognitive processes rather than support them \cite{macnamara2024does}. Framed within the broader literature on AI and skill acquisition \cite{rafner2021deskilling, natali2025ai}, our findings are consistent with upskilling inhibition, where reliance on automated support limits the development of underlying competencies. 

Our results add nuance by showing that AI’s impact on learning might depend not only on access, but also on how users choose to engage with AI and the informativeness of the assistance, consistent with evidence that cognitively engaged AI use, rather than passive delegation, preserves learning outcomes \cite{shen2026ai}. In our setting, a limited amount of information fails to improve immediate performance while still interrupting independent reasoning, thereby harming subsequent learning. This pattern suggests that AI assistance with low information value may end up being more distracting than helpful, drawing attention away from the effort that could be spent on independent learning. In contrast, more informative AI assistance can potentially support skill development, especially for users who engage with it selectively and strategically. For individuals with lower initial ability, however, access alone may be insufficient and potentially counterproductive without the type of additional guidance, monitoring, or incentive mechanisms that encourage sustained cognitive engagement. Taken together, our results highlight the importance of AI usage design and scaffolding, consistent with emerging work on AI systems that promote user engagement and learning \cite{ma2026learning, tran2026pacing}, suggesting that effective AI deployment in learning contexts requires careful attention to system capability, user heterogeneity, and usage design.

Our findings should be interpreted within the study’s scope. 
First, we focus on short-term effects after limited exposure to AI, rather than long-term skill development. It remains unclear whether these patterns persist or change with sustained use, which calls for future longitudinal work. Second, while our controlled, logic-based task allows for precise measurement, it differs from real-world settings that often involve open-ended problems and conversational AI. Future research should examine these dynamics in more high-stakes and domain-specific contexts to better understand how well they generalize.

\section{Conclusion}
In conclusion, our findings highlight a key trade-off in AI-assisted problem-solving: while AI can boost short-term performance, heavy reliance can reduce independent effort and hinder learning outcomes, especially for lower-ability individuals. These effects are influenced by the type and amount of assistance provided. Overall, the impact of AI on human learning depends not just on access, but on how it is used. The results suggest, in general, that we need to pay closer attention to AI’s effects on cognitive effort, reliance patterns, and long-term learning, as well as to its impact in complex, high-stakes, and specialized fields.

\section*{Acknowledgments}
We thank the reviewers for their helpful suggestions in improving the paper.  This work was supported by National Science Foundation under award NSF 2505006, by the Hasso Plattner Institute (HPI) Research Center in Machine Learning and Data Science at UCI, and by funding support from Google and from SAP.

\newpage
\bibliography{ref.bib}

\newpage

\end{document}